%% file: PaperForSubmission.tex
\documentclass[10pt,twocolumn,letterpaper]{article}

\usepackage[pagenumbers]{cvpr} 

\usepackage{graphicx}
\usepackage{amsmath}
\usepackage{amssymb}
\usepackage{booktabs}

\usepackage[pagebackref,breaklinks,colorlinks]{hyperref}

\usepackage[capitalize]{cleveref}
\crefname{section}{Sec.}{Secs.}
\Crefname{section}{Section}{Sections}
\Crefname{table}{Table}{Tables}
\crefname{table}{Tab.}{Tabs.}

\def\cvprPaperID{18} 
\def\confName{CVPR}
\def\confYear{2023}

\begin{document}

\title{Self-supervised Interest Point Detection and Description for Fisheye and Perspective Images}
\author{Marcela Mera-Trujillo\thanks{Denotes equal contribution.} \quad
Shivang Patel$^*$ \quad
Yu Gu \quad
Gianfranco Doretto\\
West Virginia University\\
Morgantown, WV 26506\\
{\tt\small \{mameratrujillo, sap00008, yugu, gidoretto\}@mix.wvu.edu}
}
\maketitle


\input{tex/0-Abstract.tex}

\input{tex/1-Introduction.tex}
\input{tex/2-RelatedWork.tex}
\input{tex/3-Description.tex}
\input{tex/4-Experiments.tex}
\input{tex/5-Conclusion.tex}

\input{tex/6-Acknowledgments.tex}

{\small
\bibliographystyle{ieee_fullname}
\bibliography{egbib}
}

\end{document}

%% file: tex/0-Abstract.tex
\begin{abstract}

  Keypoint detection and matching is a fundamental task in many computer vision problems, from shape reconstruction, to structure from motion, to AR/VR applications and robotics. It is a well-studied problem with remarkable successes such as SIFT, and more recent deep learning approaches. While great robustness is exhibited by these techniques with respect to noise, illumination variation, and rigid motion transformations, less attention has been placed on image distortion sensitivity. In this work, we focus on the case when this is caused by the geometry of the cameras used for image acquisition, and consider the keypoint detection and matching problem between the hybrid scenario of a fisheye and a projective image. We build on a state-of-the-art approach and derive a self-supervised procedure that enables training an interest point detector and descriptor network. We also collected two new datasets for additional training and testing in this unexplored scenario, and we demonstrate that current approaches are suboptimal because they are designed to work in traditional projective conditions, while the proposed approach turns out to be the most effective.
 \end{abstract}
 

%% file: tex/1-Introduction.tex
\section{Introduction}
\label{sec:introduction}

Many computer vision tasks require the automated detection of corresponding keypoints in multiple images, ranging from Structure-from-Motion~\cite{schonberger2016structure, schonberger2015single} and 3D reconstruction~\cite{mohr1995relative, schonberger2015single} to SLAM~\cite{mur2015orb, engel2014lsd}, AR/VR~\cite{wang2020vr} applications, and many others. The typical approach is to use an algorithm that we call interest point detector and descriptor. It addresses the problem of detecting keypoints and assigns a descriptor to them for matching across images. This is a very well studied area, and the popular solutions are based on classic algorithms like SIFT~\cite{lowe2004distinctive}, or the most recent deep learning based approaches like SuperPoint~\cite{detone2018superpoint}.

Despite the recent advances in this field, the robustness against image distortions has received less attention compared to other factors such as noise, illumination variation, and rigid motion transformations in terms of the common metrics of repeatability and matching. This issue is crucial when dealing with keypoint matching between images captured by cameras with different geometries, as in the case of a rover equipped with a fisheye camera doing visual odometry over terrain previously mapped with a perspective camera. The hybrid camera scenario just described is challenging because keypoints may easily undergo image distortions that can hinder their detection or matching accuracy.

In this work, we set out to address the challenge of keypoint detection and matching in the hybrid camera scenario. We specifically focus on the case of fisheye images and perspective images. We build on a popular approach like~\cite{detone2018superpoint}, and we show that by deriving the hybrid homography that relates fisheye and projective images, it is possible to design a self-supervised training procedure that is effective for this particular case. We also design a new set of losses and show that recent contrastive losses used for self-supervised learning are effective for descriptor matching. In order to train and test our approach, we also developed two datasets, which we plan to release to the public. One is made of synthetic images generated from a video game, and the other was collected with real cameras. We tested the approach also on previously available datasets, but the data variety of current benchmarks in this domain is limited to do a robust training. The results demonstrate that the proposed approach is showing good promise by exhibiting the state-of-the-art metrics in the hybrid camera scenario while highlighting the difficulties of the current approaches that focus on the traditional perspective case.


%% file: tex/2-RelatedWork.tex
\section{Related Work}
\label{sec:rel_work}

Existing feature extraction and descriptor methods can be categorized into two groups: traditional approaches and learning-based approaches.

\textbf{Traditional approaches.} 
Over the years, several approaches have been developed to efficiently solve the point feature extraction problem. One of the oldest and most basic feature detection approaches is the Harris Corner Detector \cite{harris1988combined}, which distinguishes between edges and corners. Another method, SIFT \cite{lowe2004distinctive}, is scale-invariant and aims to solve issues related to intensity, viewpoint changes, and image rotation in feature matching. To address the slowness of SIFT, the SURF \cite{bay2006surf} algorithm was developed, which is faster and more efficient than SIFT, while still being robust and exhibiting similar matching performance. Another alternative to SIFT and SURF is FAST \cite{rosten2006machine}, which is a faster corner detection algorithm used for real-time applications. Additionally, BRIEF \cite{calonder2010brief} can work with any other feature detector since it does not provide any method to find the features; it converts any other feature descriptors in floating-point numbers to binary strings. ORB \cite{rublee2011orb} and AKAZE \cite{akaze} are efficient alternatives to SIFT or SURF that use the FAST keypoint detector and BRIEF descriptor, providing more efficient performance.

While the majority of point feature detection and descriptor models work well in perspective images, they are less effective in fisheye images \cite{sagetong}. Some works have addressed this problem by either modifying the previous methods to work on omnidirectional images \cite{arican2010omnisift}, or by transforming the omnidirectional image to a perspective image and then applying traditional methods \cite{ma20153d}. However, our paper takes a different approach by using direct fisheye and perspective images as input, rather than altering the nature of the data to address the feature extraction problem.

\textbf{Learning-based approaches.} 
Feature extraction and descriptor techniques typically involve three steps: detecting keypoints, estimating orientation, and extracting robust descriptors. Initially, works such as TILDE \cite{verdie2015tilde}, \cite{lenc2016learning}, \cite{yi2016learning}, and \cite{simo2015discriminative} successfully tackled each of these problems individually. Later, LIFT \cite{yi2016lift} proposed a deep network architecture that unified the three steps by detecting keypoints, estimating orientation, and extracting feature descriptors in a single process. This pattern of unified approaches can be seen in subsequent methods. For example, SuperPoint \cite{detone2018superpoint} is a self-supervised framework that computes both keypoints and descriptors, but it performs poorly on rotation. In contrast, LF-Net \cite{ono2018lf} uses a two-branch network setup to learn sparse keypoints and descriptors that are both scale invariant and oriented. Similarly, D2-net \cite{dusmanu2019d2} and R2D2 \cite{revaud2019r2d2} focus on reliable dense feature descriptors and feature detectors. More recently, RoRD \cite{parihar2021rord} proposed a framework that extends D2-net by learning rotation-robust local descriptors through data augmentation and orthographic viewpoint projection. Moreover, FisheyeSuperPoint~\cite{konrad2021fisheyesuperpoint} fine-tunes SuperPoint to improve keypoint detection on fisheye images.

Despite the significant progress in learning based frameworks, no other previous works addresses the learning based feature extraction and descriptor in a hybrid scenario. To the best of our knowledge, \cite{bastanlar2012multi, lin2014hopis} are the methods closest to ours that in some way address the hybrid scenario, although they do so by taking a traditional approach.


%% file: tex/3-Description.tex
\section{Hybrid Interest Point Model}
\label{sec:description}

Our goal is to design an interest point detector and descriptor that is effective in a hybrid camera system scenario, meaning that pairs of corresponding keypoints to be selected and matched, can belong to images acquired by cameras with different geometries. We specifically focus on the hybrid case of fisheye and perspective camera models.

\subsection{Architecture}
\label{sec-architecture}

\begin{figure}[t]
  \centering
  \includegraphics[width=\linewidth]{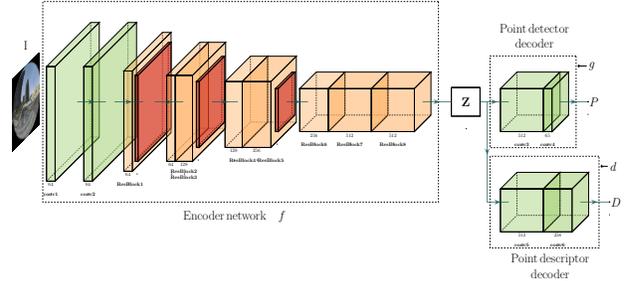}
  \caption{\textbf{Hybrid Interest Point Architecture.} Architecture of the hybrid interest point descriptor and detector.}
  \label{fig:hybridmodel}
  \vspace{-3mm}
\end{figure}

The architecture and methods build on the SuperPoint approach~\cite{detone2018superpoint}, with some notable differences. It is summarized in Figure~\ref{fig:hybridmodel}.
It comprises of three major components. Given an input image $I$, of size $h \times w$, a shared \emph{encoder} network $f$ produces a feature $\mathbf{Z} = f(I)$. $\mathbf{Z}$ is the input of a \emph{point detector decoder} network $g$ that generates a pixel-wise map $\mathcal{P} = g(\mathbf{Z})$, where $\mathcal{P}_i$ indicates the probability of an interest point being located at pixel $i$. $\mathbf{Z}$ is also the input of a \emph{point descriptor decoder} network $d$ that generates a pixel-wise map $\mathcal{D} = d(\mathbf{Z})$, where $\mathcal{D}_i$ is the descriptor of pixel $i$, which has length 256, and $\ell_2$-norm one.

Compared to~\cite{detone2018superpoint} the encoder network $f$ is a ResNet-18~\cite{he2016deep} with some modifications. We retain only the 18 convolutional layers, and we insert three $2 \times 2$ non-overlapping max-pooling layers, after the 5-th, 10-th, and 14-th convolutional layers. Therefore, $\mathbf{Z}$ has dimensions ${h_c \times w_c \times 512}$, where $h_c = h/8$, and $w_c = w/8$. In this way, the decoders $g$ and $d$ are the same as in~\cite{detone2018superpoint}, only that their first convolutional layers accept an input feature with 512 channels, rather than 256.  

The architecture is fully convolutional, and the intent is to process the whole input image, which could be either a fisheye or a perspective image, with a single feed-forward computation to generate interest point detections and descriptors. The detections are produced by the pipeline $g \circ f$, and the descriptors by the pipeline $d  \circ f$, so that the computations made by $f$ are shared.

\subsection{Loss Functions}

The three networks $f$, $g$, and $d$ are trained jointly, with a self-supervised procedure. The point detection pipeline $g\circ f$ is trained in a supervised manner, where for an image $I$ we use a set of pixel-wise automatically generated pseudo-labels $\mathcal{Y}$ indicating interest point positions. In \S~\ref{sec-training} we explain how $\mathcal{Y}$ is obtained.

The detection loss function used is designed around the architecture described in \S~\ref{sec-architecture}. In particular, the image $I$ is virtually divided in $h_c \times w_c$ disjoint cells of size $8 \times 8$ pixels, and the output $\mathcal{P}_{i,j}$ of a cell in position $(i,j)$ is produced by a softmax layer, which has 65 outputs to handle the case when there are no interest points present. So, if $\mathcal{Y}_{i,j}$ indicates the labels of cell $(i,j)$, then the detection loss function for image $I$ is
\begin{equation}
  \mathcal{L}_{g\circ f} ( I, \mathcal{Y} ) = \frac{1}{h_c w_c} \sum_{i,j = (1,1)}^{(h_c,w_c)} \ell(\mathcal{P}_{i,j}, \mathcal{Y}_{i,j}) \;,
  \label{eq-det-crossentropy}
\end{equation}
where $\ell (\cdot)$ is the cross-entropy loss. Note that if there is more than one interest point in cell $(i,j)$, only one is randomly picked. 

For a hybrid interest point detector it is necessary to detect points in fisheye as well as perspective images. Typically, fisheye images have a significantly larger field of view than perspective images, and it is reasonable to assume that they will contain more image structures, leading to more interest points. Therefore, for a given fisheye image $I$ we assume that there will be $K$ perspective images $I_1', \ldots , I_K'$ with a field of view that significantly overlaps with the one of image $I$. In \S~\ref{sec-generation} we describe our approach to randomly generate the set $\{I_k'\}$ automatically from $I$, with field of view that fully overlaps with the one of $I$. Also, from $\{I_k'\}$ we can generate the pseudo-labels $\{ \mathcal{Y}_k'\}$. In this way, the complete \emph{detection loss} for image $I$ becomes
\begin{equation}
  \small
\mathcal{L}_{det} ( I, \{I_k'\}, \mathcal{Y}, \{\mathcal{Y}_k' \} ) = \mathcal{L}_{g\circ f} ( I, \mathcal{Y}) + \frac{1}{K} \sum_{k = 1}^{K} \mathcal{L}_{g\circ f} ( I_k', \mathcal{Y}_k') \; ,
\end{equation}

To learn instead that corresponding points in a fisheye image $I$ and a perspective image  $I_k'$ should have the same descriptor, we set up a contrastive prediction task~\cite{chen2020simple}. Specifically, given the architecture of the descriptor decoder $d$, similarly to the detector decoder $g$, an image is divided in the same set of $8 \times 8$ cells, and $d \circ f$ predicts the descriptor for each cell. The pixel-wise descriptors are obtained via bicubic interpolation (see~\cite{detone2018superpoint} for details). Therefore, let $\mathbf{d} \in \mathcal{D}$ be a cell descriptor of image $I$. Let us also assume that $\mathcal{H}_k$ is the image domain transformation that has mapped $I$ onto $I_k$ (which will be defined in \S~\ref{sec-generation}). Then, the centroid position of the cell of $\mathbf{d}$ will be potentially mapped, according to $\mathcal{H}_k$, onto a cell of $I_k'$. Let us indicate the descriptor of that cell with $\mathbf{d}'_{k,\mathbf{d}}$. We expect $\mathbf{d}$ and $\mathbf{d}'_{k,\mathbf{d}}$ to be as close as possible since they describe the same region in the two images. So, $(\mathbf{d},  \mathbf{d}'_{k,\mathbf{d}})$ will form a positive pair. The other pairs  $(\mathbf{d},  \mathbf{d}')$, where $\mathbf{d}' \in \mathcal{D}_k'$ does not represent the cell of $ \mathbf{d}'_{k,\mathbf{d}}$, and are expected to be made of different descriptors, and thus are negative pairs. Therefore, the descriptor loss for the pair of images $I$, and $I_k'$, is given by
\begin{equation}
  \small
  \mathcal{L}_{d \circ f}(I, I_k', \mathcal{H}_k) = \frac{1}{|\mathcal{D}_k|} \sum_{ \mathbf{d} \in \mathcal{D}_k } - \log \frac{  \exp{ (\mathbf{d}^{\top} \mathbf{d}'_{k,\mathbf{d}} / \tau )} }{ \sum_{\mathbf{d}' \in \mathcal{D}_k'} \exp{ (\mathbf{d}^{\top} \mathbf{d}' / \tau ) } } \; ,
  \label{eq-descriptor}
\end{equation}
where $\mathcal{D}_k \subseteq  \mathcal{D}$ is the subset of descriptors with centroid location that maps onto a location inside image $I_k'$. The complete \emph{descriptor loss} function for image $I$ is 
\begin{equation}
\mathcal{L}_{des}(I, \{I_k'\}, \{\mathcal{H}_k\}) = \frac{1}{K} \sum_{k=1}^K \mathcal{L}_{d \circ f}(I, I_k', \mathcal{H}_k) \; .
\end{equation}
Finally, the \emph{total loss} for the joint training on a per fisheye image basis is 
\begin{equation}
 \mathcal{L} =  \mathcal{L}_{det} + \gamma \mathcal{L}_{des} \; ,
  \label{eq:Loss}
\end{equation}
where $\gamma$ strikes a balance between the detection and descriptor losses.

\subsection{Training Procedure}
\label{sec-training}

\begin{figure}[t]
  \centering
  \includegraphics[width=0.242\linewidth]{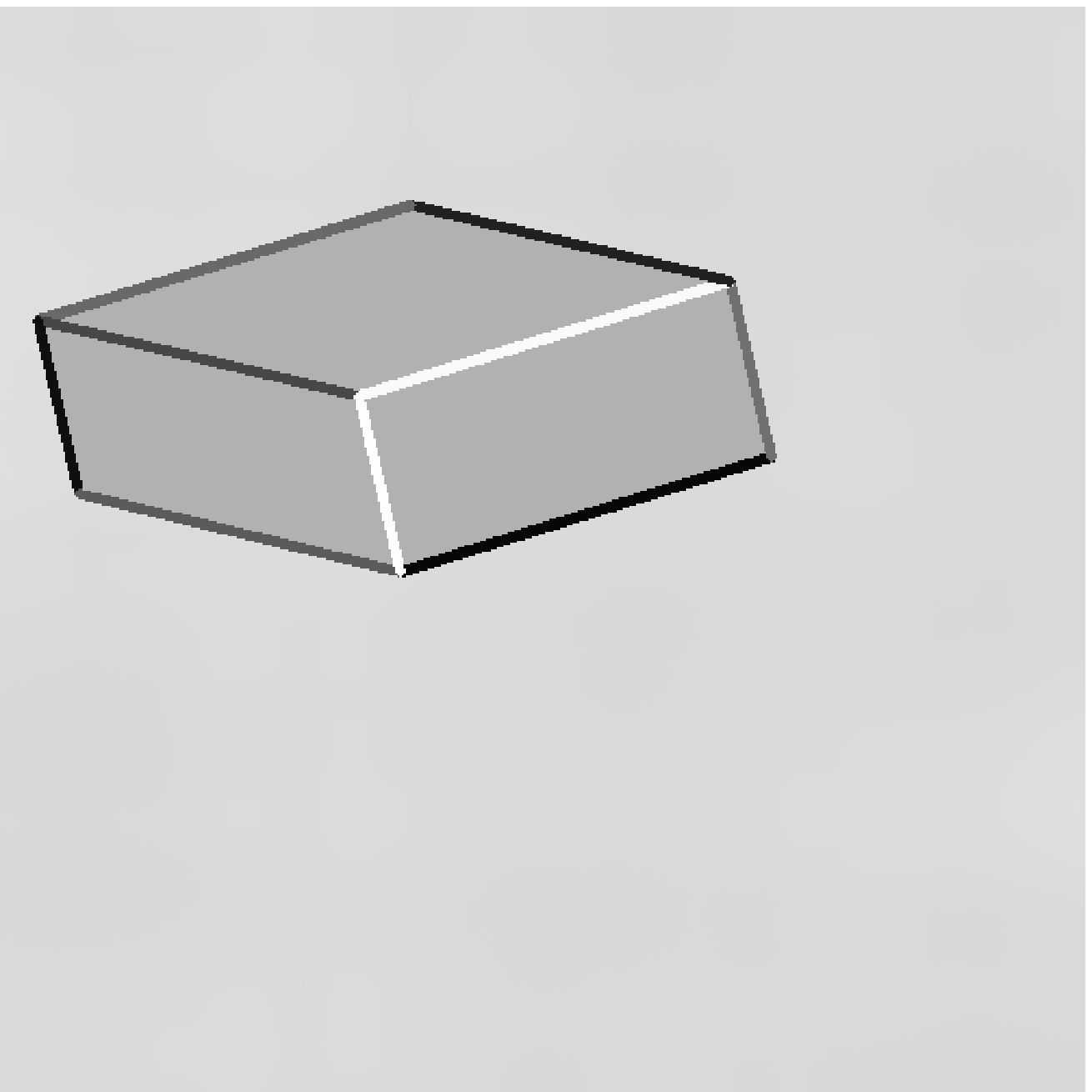}
  \includegraphics[width=0.242\linewidth]{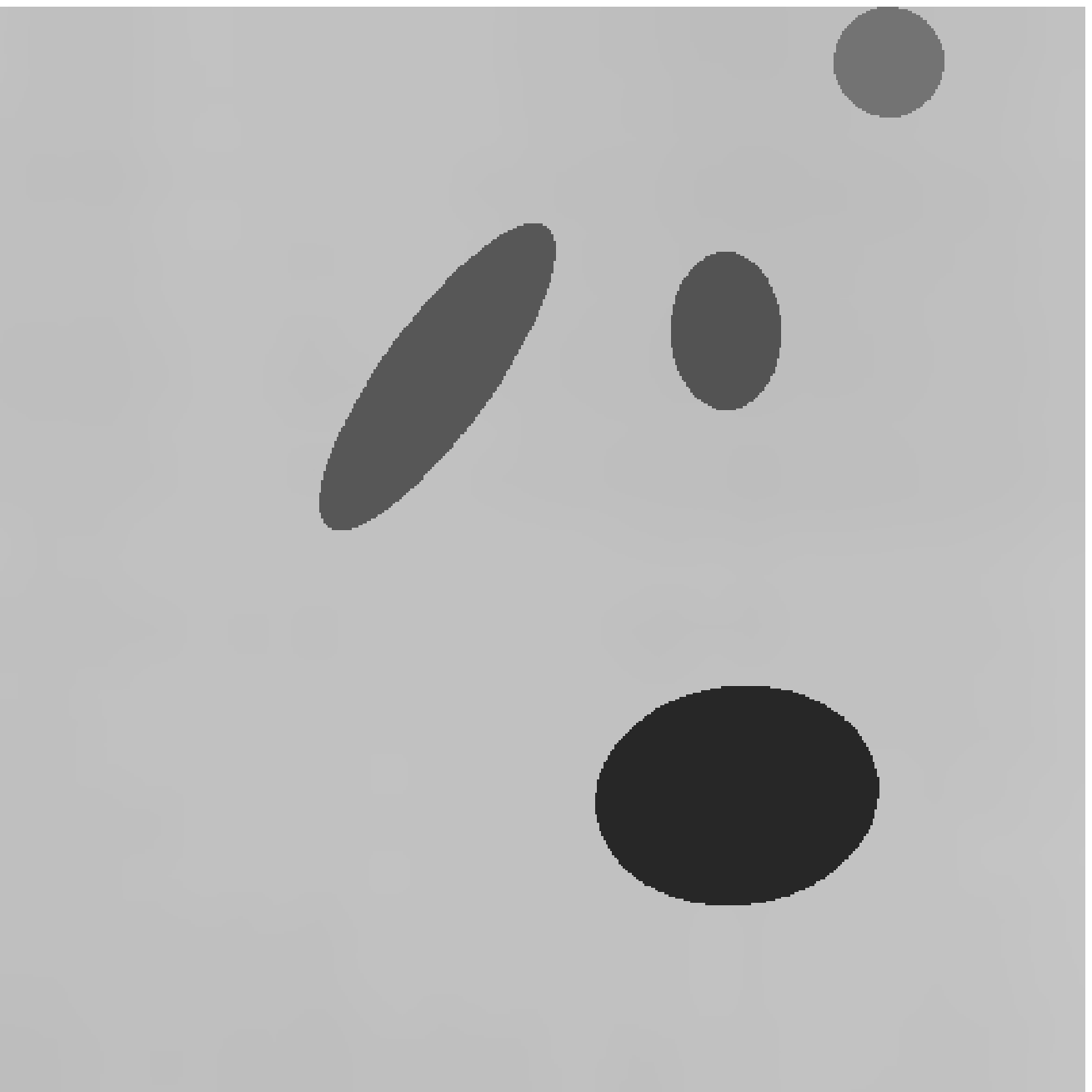}
  \includegraphics[width=0.242\linewidth]{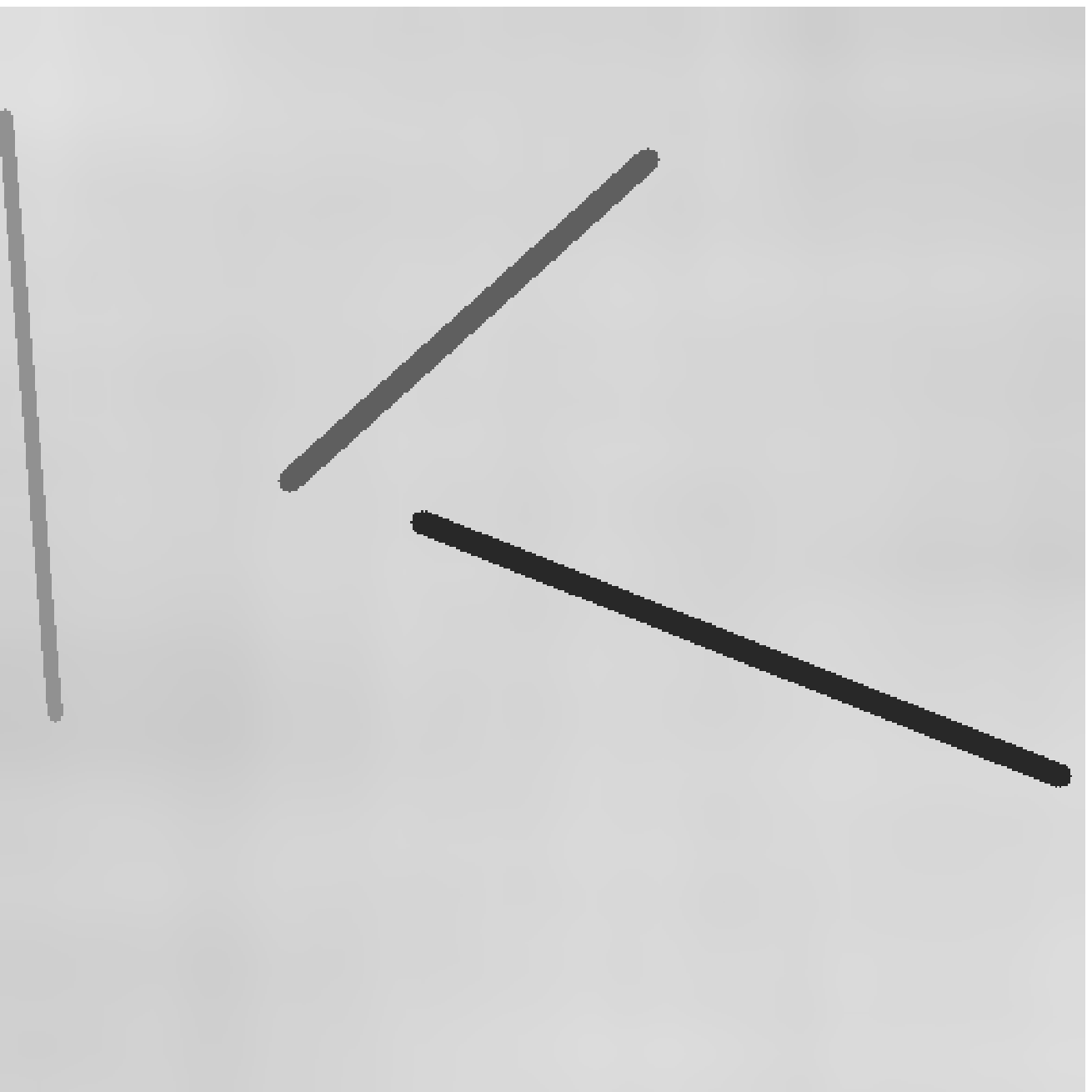}
  \includegraphics[width=0.242\linewidth]{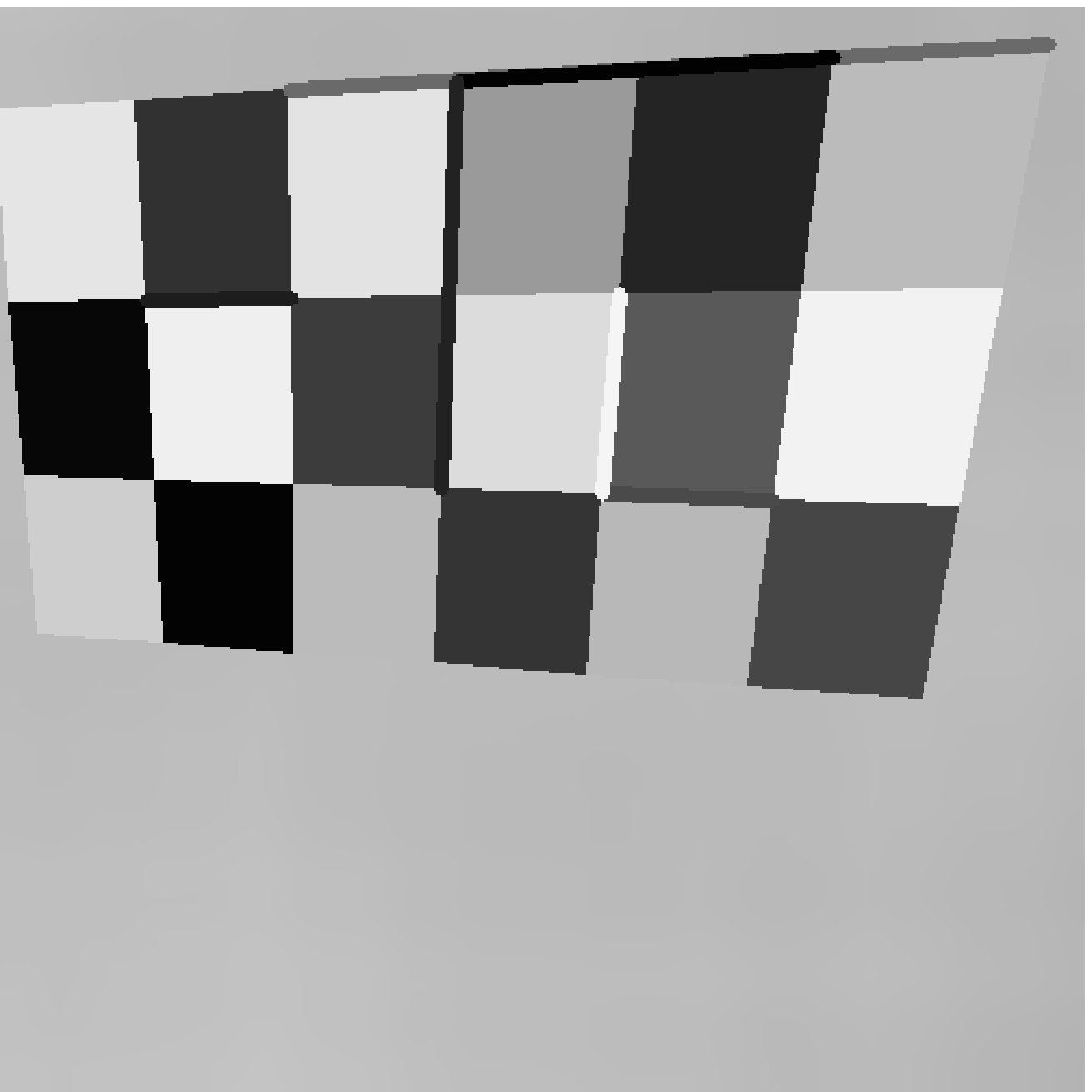}
  \caption{\textbf{Synthetic images with geometric primitives.} Image samples of size $320 \times 320$ created at runtime to pre-train the interest point detector pipeline.}
  \label{fig:kornia-images}
    \vspace{-3mm}
\end{figure}

The first training step is to arrive at a mechanism for generating pseudo-labels $\mathcal{Y}$ from the image $I$. Similar to~\cite{detone2018superpoint}, we train the detector pipeline $g \circ f$ from scratch with synthetically generated images like triangles, lines, polygons and so on, which have 2D geometric primitives that allow to identify interest points unambiguously, like corresponding to T-junctions, Y-junctions, L-junctions, or center of elipsoidal blobs. We use OpenCV and the Kornia image library to generate 2.4M synthetic images with relative interest point annotations that we use for pre-training the detector with loss~\eqref{eq-det-crossentropy} in a self-supervised manner. The images are randomly generated and are seen only once by the detector during training using batches of size 16. The training continues for 150K iterations, processing the same number of batches. Figure~\ref{fig:kornia-images} shows samples of such images. 

The next training step is to use another self-supervised process called Homographic Adaptation, described in~\cite{detone2018superpoint}. The goal is to make the detector pipeline work well in more general scenarios, and also with real images. We use Homographic Adaptation where we generate random homographies as explained in \S~\ref{sec-random-homography}, and we leverage the perspective training images of the KITTI-360 dataset~\cite{Liao2021ARXIV}, together with synthetic perspective training images that we generate from the game Grand Theft Auto V (GTAV), as explained in \S~\ref{subsec:SyncData}, for an enhanced exposure to image variety. We used 12K images from KITTI-360, 30K images from GTAV, all resized to $320\times320$, and trained for more than 100K iterations until convergence, forming batches of 16 images, and using loss~\eqref{eq-det-crossentropy}. For interest point superset generation we sampled 100 random homographies per image. We repeat Homographic Adaptation twice.

After completing the two steps above, given a fisheye image $I$, and corresponding perspective images $\{ I_k' \}$, we can use the current detector pipeline $g\circ f$ to compute the \emph{pseudo-labels} $\mathcal{Y}$ and $\{ \mathcal{Y}_k' \}$. At this point training of all the networks $f$, $g$, and $d$ can continue with the joint total loss \eqref{eq:Loss}. This is based on using fisheye images from the KITTI-360 dataset, and the synthetic dataset GTAV-Hybrid that we introduce in \S~\ref{subsec:SyncData}. More details are included in \S~\ref{sec:experiments}.

\section{Generation of Perspective Images}
\label{sec-generation}

Given a fisheye image $I$ we show how to produce a perspective image $I'$. This will allow to randomly sample the set $\{ I_k' \}$, as well as the set of transformations $\{ \mathcal{H}_k \}$.

\subsection{Hybrid Homography Model}

\begin{figure}[t]
  \centering
  \includegraphics[width=0.8\linewidth]{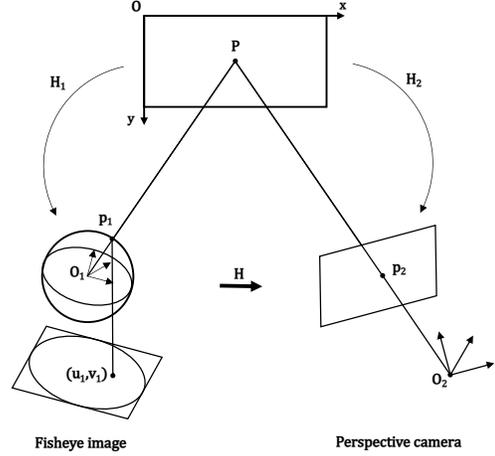}
  \caption{\textbf{Hybrid fisheye and perspective imaging system.} $O_1$ is the camera center of a fisheye camera and $O_2$ is the camera center of a perspective (conventional) camera.}
  \label{fig:hybrid}
    \vspace{-3mm}
\end{figure}

We derive a model that is a hybrid homography between a fisheye and a perspective image. With reference to Figure~\ref{fig:hybrid}, we consider a system with an omnidirectional camera with fisheye lenses, centered in $O_1$, and a perspective camera centered in $O_2$. The cameras are observing a 3D scene made by a planar surface. Let $P$ be a point on such surface, represented in homogeneous coordinates with respect to a world reference frame with the origin on the surface and the xy-plane parallel to the surface. Let $p_1$ be a 3D vector pointing towards $P$ from $O_1$, then the following relationship holds~\cite{scaramuzza2006toolbox}
\begin{equation}
  \lambda_1 p_1 = \lambda_1 \left[
    \begin{array}{c}
      u_1 \\
      v_1 \\
      \varphi (u_1,v_1)
    \end{array} 
  \right] = \Pi_1 P \; ,
\end{equation}
where $\lambda_1$ is an appropriate scalar, $(u_1, v_1)$ represent the sensor coordinates,
$\varphi (u_1,v_1)$ is a polinomial function in $\rho_1 = \sqrt{u_1^2 + v_1^2}$, and $\Pi_1 = [R_1, T_1]$ is a projection matrix, where the extrinsic parameters $(R_1, T_1)$ map $P$ onto the coordinate system of the camera. Finally, since $P = [X, Y, 0, 1]^{\top}$ because it is on the planar surface, we also have that 
\begin{equation}
  \lambda_1 p_1 = [R_1, T_1] \left[
    \begin{array}{c}
      X \\
      Y \\
      0 \\
      1
    \end{array} \right] = H_1 \left[
    \begin{array}{c}
      X \\
      Y \\
      1
    \end{array} \right] \; ,
\end{equation}
where $H_1$ is a $3 \times 3$ matrix obtained from $[R_1, T_1]$ by removing its third column.

Similarly, let $p_2$ be a 3D vector pointing towards $P$ from $O_2$, then the following relationship holds
\begin{equation}
  \lambda_2 p_2 = \lambda_2 \left[
    \begin{array}{c}
      u_2 \\
      v_2 \\
      1
    \end{array} 
  \right] = \Pi_2 P \; ,
\end{equation}
where $\lambda_2$ is a suitable parameter.
If $\Pi_2 = [R_2, T_2]$ is the projection matrix, where $(R_2, T_2)$ map $P$ onto the coordinate system of the perspective camera, it follows that 
\begin{equation}
  \lambda_2 p_2 = [R_2, T_2] \left[
    \begin{array}{c}
      X \\
      Y \\
      0 \\
      1
    \end{array} \right] = H_2 \left[
    \begin{array}{c}
      X \\
      Y \\
      1
    \end{array} \right] \; ,
\end{equation}
and $H_2$ is $[R_2, T_2]$ without its third column.

From the discussion above it follows that 
\begin{equation}
    \lambda \left[
    \begin{array}{c}
      u_2 \\
      v_2 \\
      1
    \end{array} 
  \right] = H
  \left[
    \begin{array}{c}
      u_1 \\
      v_1 \\
      \varphi (u_1,v_1)
    \end{array} 
  \right] \; ,
  \label{eq-hybrid-homography}
\end{equation}
where $\lambda$ is a suitable scalar, $H \doteq H_2 H_1^{-1}$ is a homography that encodes the geometric relationship between the planar scene and the camera system. The mapping $\mathcal{H} : (u_1, v_1) \mapsto (u_2, v_2)$ defined by~\eqref{eq-hybrid-homography} is what we refer to as the \emph{hybrid homography} of the system.

To derive a mapping between pixel coordinates, recall that $(u_1,v_1)$ is related to the fisheye image coordinates $(x_1, y_1)$ via an affine transformation $(A,t)$, and that $p_2$ is related to the perspective image coordinates $(x_2, y_2)$ via the intrinsic parameters matrix $\mathtt{K}$. This gives the relationship 
\begin{equation}
    \lambda \left[
    \begin{array}{c}
      x_2 \\
      y_2 \\
      1
    \end{array} 
  \right] = \mathtt{K} H
  \left[
    \begin{array}{c}
      A [x_1, y_1]^{\top} + t \\
      \varphi (A [x_1, y_1]^{\top} + t)
    \end{array} 
  \right] \; .
  \label{eq-hybrid-homography2}
\end{equation}

\subsection{Random Generation of Homographies}
\label{sec-random-homography}

The random generation of hybrid homographies can occur by sampling the matrix $H$. We can consider the intrinsics of the cameras to remain constant, as it often happens during the collection of a dataset. One way to sample $H$ would be to randomly sample the quadruple $(R_1, T_1, R_2, T_2)$. However, this approach is difficult to control, and we take a more direct approach.

First, we observe that when the cameras share the origin, i.e., $O_2=O_1$, and their reference systems are aligned, then $R_2 = R_1$, $T_2 = T_1$, and $H$ is the identity. We then perturb this configuration with a series of transformations. One is an in-plane 2D rotation $H_R$~\eqref{eq-h-rotation}, then a 2D scaling $H_s$~\eqref{eq-h-scaling-shear}, then a 2D skew $H_{k}$~\eqref{eq-h-scaling-shear}, then a shear $H_{h}$~\eqref{eq-h-xy-shear-translation} of the xy-plane into the plane passing through $(0,0,0)$, $(1, 0, h_x)$, and $(0,1,h_y)$, and finally a 2D translation $H_{T}$~\eqref{eq-h-xy-shear-translation}. The transformations are summarized as follows
\begin{equation}
  H_R = \begin{bmatrix}
    \cos a & \sin a & 0 \\
    -\sin a & \cos a & 0 \\
    0 & 0 & 1
    \end{bmatrix} \; ,
  \label{eq-h-rotation}
\end{equation}
\begin{equation}
  H_s = \begin{bmatrix}
      s_x & 0 & 0 \\
      0 & s_y & 0 \\
      0 & 0 & 1
      \end{bmatrix} \; , \quad 
  H_{k} = \begin{bmatrix}
        1 & k_x & 0 \\
        k_y& 1 & 0 \\
        0 & 0 & 1
        \end{bmatrix} \; ,
  \label{eq-h-scaling-shear}
\end{equation}
\begin{equation}
  H_{h} = \begin{bmatrix}
        1 & 0 & 0 \\
        0 & 1 & 0 \\
        h_x & h_y & 1
        \end{bmatrix} \; , \quad
  H_{T} = \begin{bmatrix}
        1 & 0 & t_x \\
        0 & 1 & t_y \\
        0 & 0 & 1
        \end{bmatrix} \; .
  \label{eq-h-xy-shear-translation}
\end{equation}
$H$ is then randomly sampled as  $H = H_R H_s H_k H_h H_T$. This allows to generate the homographies for the Homographic Adaptation, and also the set of hybrid homography transformations $\{ \mathcal{H}_k \}$.

\subsection{Perspective Image Synthesis}
\label{sec-persective-synthesis}

As described in \S~\ref{sec-random-homography} we can sample a hybrid homography $\mathcal{H}$ as in~\eqref{eq-hybrid-homography2}, and from a fisheye image $I$ we generate a perspective image $I'$ by using inverse warping with bilinear interpolation.
Note that since the projection center of the perspective view is close if not overlapping with the projection center of the fisheye camera, and since the scene objects are normally at a much greater distance than the distance between the projection centers, model~\eqref{eq-hybrid-homography2} is fully respectful of the 3D nature of the scene, regardless of its planarity. By setting an appropriate sampling range of parameters, it is easy to draw a set of homographies $\{ \mathcal{H}_k \}$ that give rise to a set of projective images $\{ I_k' \}$, with field of view completely overlapping with the field of view of the fisheye image $I$.  

\section{GTAV-Hybrid Synthetic Dataset}
\label{subsec:SyncData}

Due to the lack of publicly available datasets with annotations that correlate fisheye and perspective images, we design a process for generating synthetic fisheye images for which the calibration parameters are known. The scenarios from which we capture images is given by Grand Theft Auto V (GTAV), a popular role playing game in an expansive virtual city. We use a software package called G2D~\cite{doan2018g2d} to capture computer generated images of GTAV. Specifically, we set up a path in the simulator and then tour the path capturing images with the perspective virtual camera with six degrees of freedom (6DoF) at specific timestamps.

\begin{figure}[t]
  \centering
  \begin{minipage}{.48\textwidth}
    \centering
    \begin{subfigure}{0.55\textwidth}
      \centering
      \includegraphics[width=\textwidth]{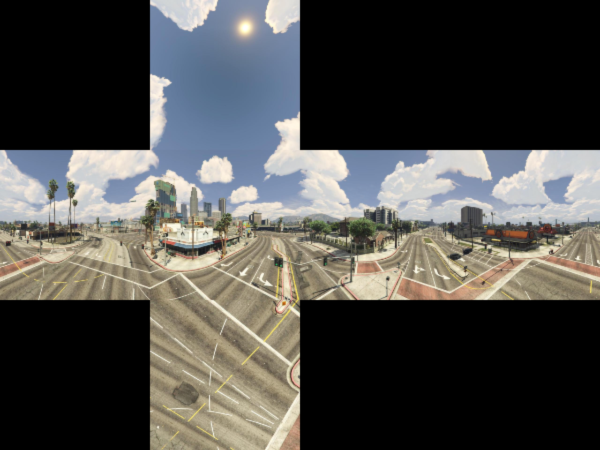}
    \end{subfigure}
    \begin{subfigure}{0.44\textwidth}
      \centering
      \includegraphics[width=\textwidth]{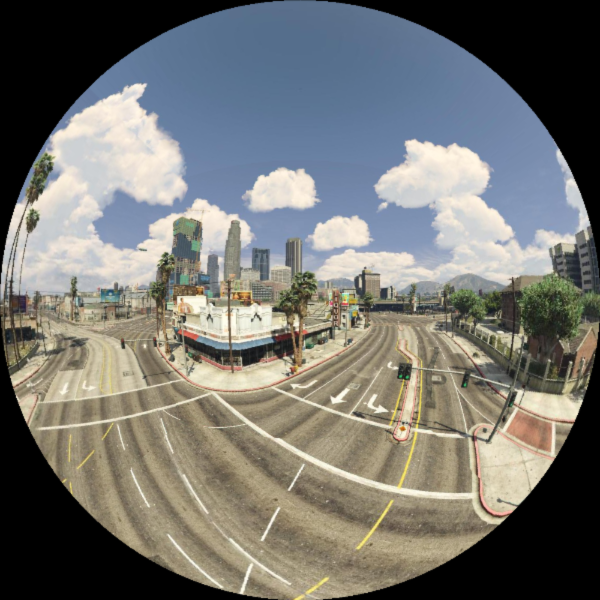}
    \end{subfigure}
  \end{minipage}
  \caption{\textbf{Cube maps and fisheye images.} Cube map generated from GTAV with G2D (left), and corresponding conversion to a fisheye image (right).}
  \label{fig:GTAVimg}
    \vspace{-3mm}
\end{figure}

Since G2D implements only a perspective virtual camera, but we are interested in capturing fisheye images, we take a \emph{cube mapping} approach. This means that at every location we capture 6 views with a virtual camera with a field of view of 90 degrees, and which points in the 6 principal directions also separated by 90 degrees. Subsequently, we follow~\cite{berenguel2020omniscv} to convert a cube map into a fisheye image. See Figure~\ref{fig:GTAVimg} for an example of a GTAV cube map and a fisheye image generated from it.

\begin{figure}[!t]
    \centering
    \begin{minipage}{.48\textwidth}
      \centering
      \begin{subfigure}{0.32\textwidth}
        \centering
        \includegraphics[width=\textwidth]{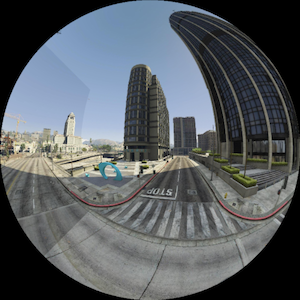}
      \end{subfigure}
      \begin{subfigure}{0.32\textwidth}
        \centering
        \includegraphics[width=\textwidth]{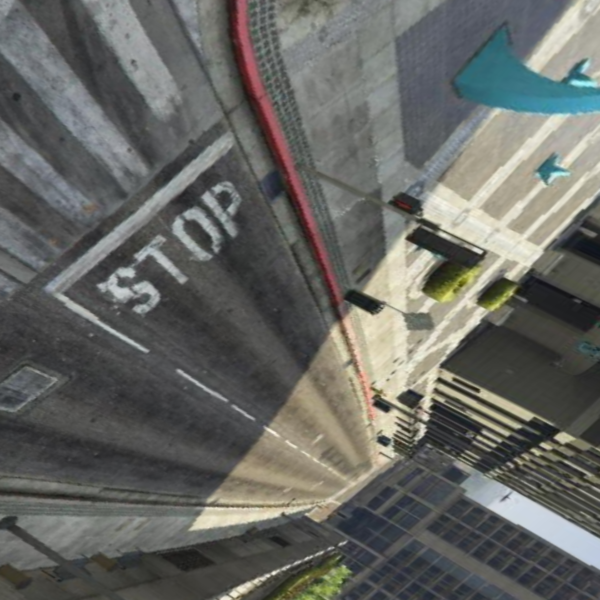}
      \end{subfigure}
      \begin{subfigure}{0.32\textwidth}
        \centering
        \includegraphics[width=\textwidth]{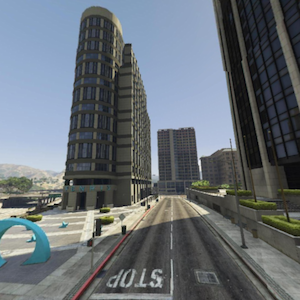}
      \end{subfigure}
      \begin{subfigure}{0.32\textwidth}
        \centering
        \includegraphics[width=\textwidth]{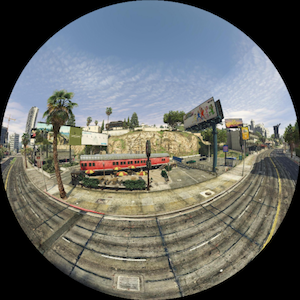}
      \end{subfigure}
      \begin{subfigure}{0.32\textwidth}
        \centering
        \includegraphics[width=\textwidth]{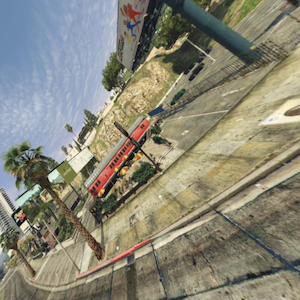}
      \end{subfigure}
      \begin{subfigure}{0.32\textwidth}
        \centering
        \includegraphics[width=\textwidth]{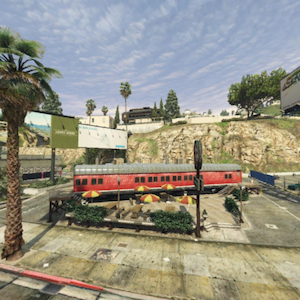}
      \end{subfigure}
      \centering
      \begin{subfigure}{0.32\textwidth}
        \centering
        \includegraphics[width=\textwidth]{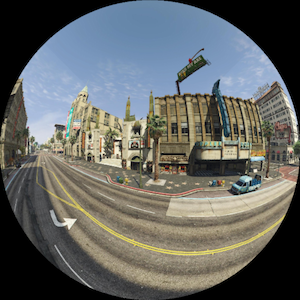}
      \end{subfigure}
      \begin{subfigure}{0.32\textwidth}
        \centering
        \includegraphics[width=\textwidth]{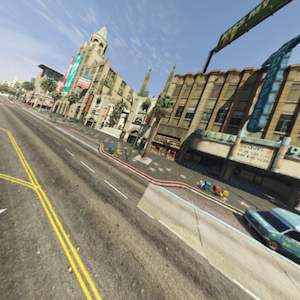}
      \end{subfigure}
      \begin{subfigure}{0.32\textwidth}
        \centering
        \includegraphics[width=\textwidth]{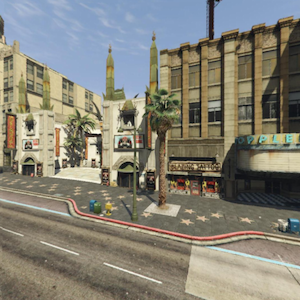}
      \end{subfigure}
    \end{minipage}
    \caption{\textbf{Paired fisheye and perspective images.} Examples of synthetic fisheye images (left), and paired perspective images (right) obtained by applying different hybrid homography transformations~\eqref{eq-hybrid-homography2}.}
    \label{fig:homography}
      \vspace{-3mm}
\end{figure}

From fisheye images, we use the procedure outlined in \S~\ref{sec-persective-synthesis} to sample perspective images. Figure~\ref{fig:homography} shows examples of fisheye images from \emph{GTAV-Hybrid}, our new dataset which will be made publicly available, and perspective images generated from them.

%% file: tex/4-Experiments.tex
\section{Experiments}
\label{sec:experiments}

\subsection{Datasets}

\textbf{GTAV-Hybrid.} We use GTAV-Hybrid for training and testing. The dataset includes a total number of 41280 fisheye images, where only 30000 fisheye images were used for training, with $K=5$ paired homography generated perspective images. We keep 600 distinct images from the remaining images for testing.

\textbf{KITTI-360.} KITTI-360~\cite{Liao2021ARXIV} comprises of a vast collection of images captured in the suburbs of Karlsruhe, Germany. Due to the sequential nature of the images, not all of them are useful for training purposes, since many of them overlap with subsequent images. To address this issue, we used a skip sampling approach to select a representative subset of 16,000 images. Out of these, 12,000 images were used for training, while the remaining 4,000 images were further reduced to a set of 600 distinct images for testing.

\textbf{Evansdale.} This is a new dataset that we have collected around our campus with a Kodak PIXPRO SP360 fisheye camera, and a Canon EOS 80D DSLR. It has 30 pairs of fisheye and perspective images. The intrinsic and extrinsic camera parameters have been extracted. This dataset will also be made publicly available.

We used GTAV-Hybrid and the KITTI-360 datasets for training and the Evansdale dataset for testing.

\subsection{Implementation Details}

The network $f$ uses ResNet-18, which includes batch normalization, max pooling, and downsampling layers. The first layer of $f$ has a kernel size of 3 instead of the original 7, followed by batch normalization and another convolutional layer with a kernel size of 3. We opted to use Leaky ReLU activations to maintain training stability throughout the network. The channel widths of the original network (64-128-256-512) were retained. To achieve a feature size of $\frac{h}{8} \times \frac{h}{8}$, where $h$ and $w$ represent the height and width of the image, we used downsampling layers.

\textbf{Training.} During the training of the interest point detector network $g\circ f$ we used a batch size of 16 and a learning rate of $10^{-3}$. Following that, we utilized the trained detector network to create pseudo ground truth labels for the GTAV-Hybrid and KITTI-360. We resized each image to $320\times 320$ and converted it to grayscale. Subsequently, we generated 100 random homographies to generate detection labels and trained the network with these images for 5,000 iterations. We repeat this Homography Adaptation step twice to achieve robust detection.

To self-supervise the hybrid point descriptor and detector model training, we required two sets of images: $K=5$ perspective images and one fisheye image. We created unique perspective images on-the-fly, given a fisheye image, throughout the training process. We keep the loss balancing term $\gamma=0.001$ and a temperature $\tau=0.15$ in~\eqref{eq-descriptor}. We used a batch size of 16 and a learning rate of $10^{-3}$, and we train for more than $100,000$ iterations until convergence.

Our method is implemented in PyTorch in a distributed setting using four Nvidia A6000 GPUs with an Intel Xenon CPU. We employed standard data augmentation techniques such as random Gaussian blur and random brightness changes in all of our training.

\begin{figure*}[t]
  \centering
  \vspace{-3pt}
  \begin{minipage}{1.\textwidth}
    \centering
    \begin{minipage}{0.33\textwidth}
      \centering
      \textbf{SIFT}
    \end{minipage}
    \begin{minipage}{0.33\textwidth}
      \centering
      \textbf{SuperPoint}
    \end{minipage}
    \begin{minipage}{0.33\textwidth}
      \centering
      \textbf{Ours}
    \end{minipage}
    \vspace{1pt}
  \end{minipage}
  \vspace{-1pt}
  \begin{minipage}{1.\textwidth}
    \centering
    \begin{subfigure}{0.33\textwidth}
      \centering
      \includegraphics[width=\textwidth]{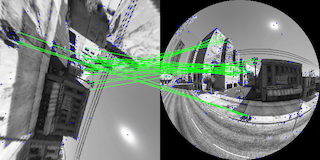}
    \end{subfigure}
    \hspace{-10pt}
    \begin{subfigure}{0.33\textwidth}
      \centering
      \includegraphics[width=\textwidth]{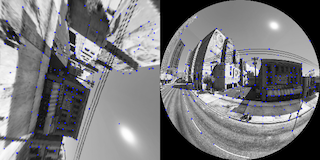}
    \end{subfigure}
    \hspace{-10pt}
    \begin{subfigure}{0.33\textwidth}
      \centering
      \includegraphics[width=\textwidth]{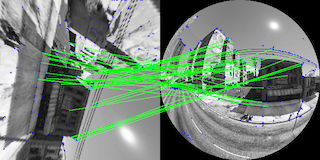}
    \end{subfigure}
  \end{minipage}
  \vspace{-1pt}
  \begin{minipage}{1.\textwidth} 
    \centering
    \begin{subfigure}{0.33\textwidth}
      \centering
      \includegraphics[width=\textwidth]{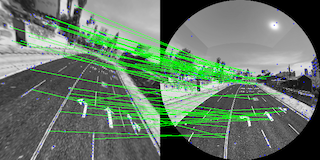}
    \end{subfigure}
    \hspace{-10pt}
    \begin{subfigure}{0.33\textwidth}
      \centering
      \includegraphics[width=\textwidth]{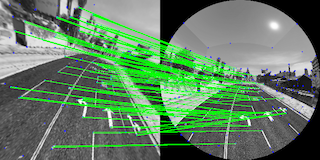}
    \end{subfigure}
    \hspace{-10pt}
    \begin{subfigure}{0.33\textwidth}
      \centering
      \includegraphics[width=\textwidth]{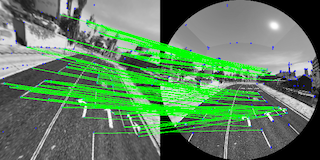}
    \end{subfigure}
  \end{minipage}
  \vspace{-1pt}
  \begin{minipage}{1.\textwidth} 
    \centering
    \begin{subfigure}{0.33\textwidth}
      \centering
      \includegraphics[width=\textwidth]{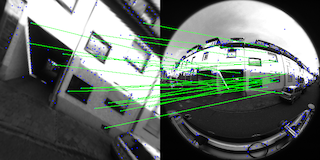}
    \end{subfigure}
    \hspace{-10pt}
    \begin{subfigure}{0.33\textwidth}
      \centering
      \includegraphics[width=\textwidth]{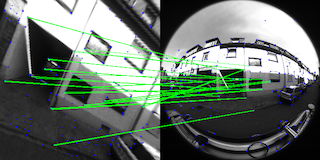}
    \end{subfigure}
    \hspace{-10pt}
    \begin{subfigure}{0.33\textwidth}
      \centering
      \includegraphics[width=\textwidth]{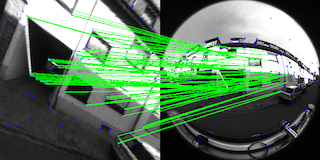}
    \end{subfigure}
  \end{minipage}
  \vspace{-1pt}
  \begin{minipage}{1.\textwidth} 
    \centering
    \begin{subfigure}{0.33\textwidth}
      \centering
      \includegraphics[width=\textwidth]{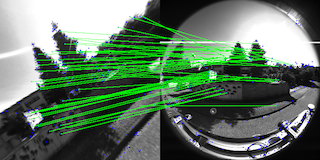}
    \end{subfigure}
    \hspace{-10pt}
    \begin{subfigure}{0.33\textwidth}
      \centering
      \includegraphics[width=\textwidth]{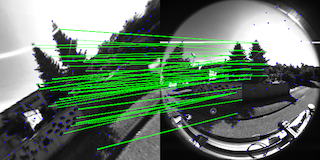}
    \end{subfigure}
    \hspace{-10pt}
    \begin{subfigure}{0.33\textwidth}
      \centering
      \includegraphics[width=\textwidth]{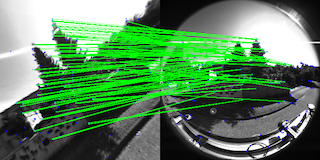}
    \end{subfigure}
  \end{minipage}
  \vspace{-1pt}
  \begin{minipage}{1.\textwidth} 
    \centering
    \begin{subfigure}{0.33\textwidth}
      \centering
      \includegraphics[width=\textwidth]{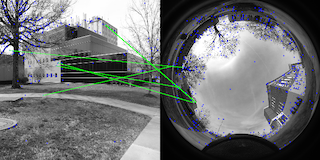}
    \end{subfigure}
    \hspace{-10pt}
    \begin{subfigure}{0.33\textwidth}
      \centering
      \includegraphics[width=\textwidth]{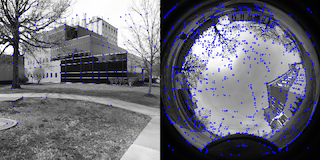}
    \end{subfigure}
    \hspace{-10pt}
    \begin{subfigure}{0.33\textwidth}
      \centering
      \includegraphics[width=\textwidth]{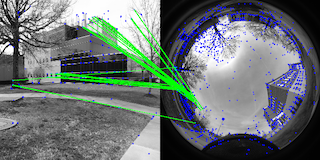}
    \end{subfigure}
  \end{minipage}
  \vspace{-1pt}
  \begin{minipage}{1.\textwidth} 
    \centering
    \begin{subfigure}{0.33\textwidth}
      \centering
      \includegraphics[width=\textwidth]{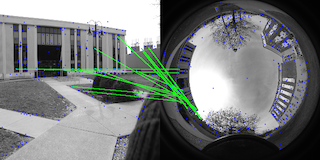}
    \end{subfigure}
    \hspace{-10pt}
    \begin{subfigure}{0.33\textwidth}
      \centering
      \includegraphics[width=\textwidth]{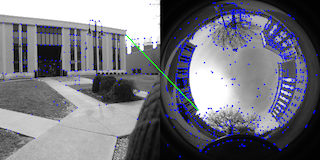}
    \end{subfigure}
    \hspace{-10pt}
    \begin{subfigure}{0.33\textwidth}
      \centering
      \includegraphics[width=\textwidth]{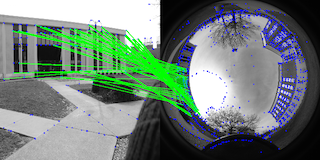}
    \end{subfigure}
  \end{minipage}
  \caption{\textbf{Qualitative Results.} First two rows are from the GTAV-Hybrid dataset, the second two from the KITTI-360 dataset, and the last two from the Evansdale dataset. The green lines indicate accurate matches. Our approach appears to perform better in cases of distortion, and is comparable to other methods in cases of minimal distortion. It's worth mentioning that SuperPoint, due to its lack of rotation invariance, encounters difficulties generating matches when confronted with rotational homography in rows 1. Similarly, it struggles to produce matches in the Evansdale dataset in rows 5 and 6 for the same reason.}\label{fig:matching}
    \vspace{-3mm}

\end{figure*}

\begin{table}
  \centering
  \scalebox{0.8}{
    \begin{tabular}{l@{\hskip 0.1in}c@{\hskip 0.1in}c@{\hskip 0.1in}c@{\hskip 0.1in}c@{\hskip 0.1in}c}
    \toprule
    & \multicolumn{2}{c}{Mean Matching} & Average number & \multicolumn{2}{c}{Repeatability}\\
    & \multicolumn{2}{c}{Score} & of matches & \multicolumn{2}{c}{Score} \\
    \midrule
     & $\epsilon = 3$ & $\epsilon = 5$ & & $\epsilon = 3$ & $\epsilon = 5$\\
    \midrule
    SIFT~\cite{lowe2004distinctive} & 0.437& 0.475 & 107 & 0.287 & 0.539 \\
    ORB~\cite{rublee2011orb} & 0.446& 0.503 & 92 & \textbf{0.488} & \textbf{0.670} \\
    AKAZE~\cite{akaze} & 0.334& 0.393 & 91.4 & 0.356 & 0.584 \\
    BRISK~\cite{6126542} & 0.185& 0.223 & 90 & 0.195 & 0.294 \\
    SuperPoint~\cite{detone2018superpoint} & 0.235 & 0.244 & 24 & 0.395 & 0.564 \\
    D2-Net~\cite{dusmanu2019d2} & 0.026 & 0.052& 18.5 & 0.207 & 0.466 \\
    RoRD~\cite{parihar2021rord} & 0.101 & 0.199& 45.42 & 0.275 & 0.526 \\
    \textbf{Ours} & \textbf{0.480} & \textbf{0.520} & 109.28 & 0.449 & 0.573\\
    \bottomrule
  \end{tabular}
  }
  \caption{Results on the GTAV-Hybrid test dataset.}
  \label{tab:gtavtest-test}
 \end{table}

 \begin{table}
  \centering
  \scalebox{0.8}{
    \begin{tabular}{l@{\hskip 0.1in}c@{\hskip 0.1in}c@{\hskip 0.1in}c@{\hskip 0.1in}c@{\hskip 0.1in}c}
    \toprule
    & \multicolumn{2}{c}{Mean Matching} & Average number & \multicolumn{2}{c}{Repeatability}\\
    & \multicolumn{2}{c}{Score} & of matches & \multicolumn{2}{c}{Score} \\
    \midrule
     & $\epsilon = 3$ & $\epsilon = 5$ & & $\epsilon = 3$ & $\epsilon = 5$\\
    \midrule
    SIFT~\cite{lowe2004distinctive} & 0.374& 0.409 & 104.53 & 0.339 & 0.542\\
    ORB~\cite{rublee2011orb} & 0.415& 0.461 & 84.70 & 0.514 & \textbf{0.633}\\
    AKAZE~\cite{akaze} & 0.325& 0.380 & 81.65 & 0.455 & 0.597\\
    BRISK~\cite{6126542} & 0.273& 0.313 & 83.32 & 0.320 & 0.461\\
    SuperPoint~\cite{detone2018superpoint} & 0.244 & 0.254 & 30.13 & \textbf{0.522} & 0.541\\
    D2-Net~\cite{dusmanu2019d2} & 0.033 & 0.074 & 16.45 & 0.257 & 0.482\\
    RoRD~\cite{parihar2021rord} & 0.113 & 0.227 & 58.0 & 0.339 & 0.544\\
    \textbf{Ours} & \textbf{0.478} & \textbf{0.518} & 172.92 & 0.455 & 0.574\\
    \bottomrule
  \end{tabular}
  }
  \caption{Results on the KITTI-360 test dataset.}
  \label{tab:kittitest-test}
 \end{table}

\begin{table}
  \centering
  \scalebox{0.8}{
    \begin{tabular}{l@{\hskip 0.1in}c@{\hskip 0.1in}c@{\hskip 0.1in}c@{\hskip 0.1in}c@{\hskip 0.1in}c}
    \toprule
    & \multicolumn{2}{c}{Mean Matching} & Average number & \multicolumn{2}{c}{Repeatability}\\
    & \multicolumn{2}{c}{Score} & of matches & \multicolumn{2}{c}{Score}\\
    \midrule
     & $\epsilon = 5$ & $\epsilon = 10$ &  & $\epsilon = 5$ & $\epsilon = 10$\\
    \midrule
    SIFT~\cite{lowe2004distinctive} & 0.043 & 0.199 & 77.16 & 0.195 & 0.332\\
    ORB~\cite{rublee2011orb} & 0.0962& 0.186 & 94.5 & 0.260 & 0.398\\
    AKAZE~\cite{akaze} & 0.152& 0.235 & 97.66 & 0.294 & 0.492\\
    BRISK~\cite{6126542} & 0.001 & 0.006  & 38 & 0.120 &0.188\\
    SuperPoint~\cite{detone2018superpoint} & 0.034 & 0.045 & 5.13 & 0.336 & \textbf{0.521}\\
    D2-Net~\cite{dusmanu2019d2} & 0.0 & 0.0 & 0.0 & 0.187 & 0.458\\
    RoRD~\cite{parihar2021rord} & 0.044 & 0.148 & 1.06 & 0.193 & 0.454\\
    \textbf{Ours} & \textbf{0.223} & \textbf{0.340} & 33.43 & \textbf{0.343} & 0.497\\
    \bottomrule
  \end{tabular}
  }
  \caption{Results on the Evansdale dataset.}
  \label{tab:evansdale-test}
\end{table}

\subsection{Results}

We evaluate our hybrid interest point detector on three datasets: GTAV-Hybrid, KITTI-360~\cite{Liao2021ARXIV}, and Evansdale. We compare our approach to state-of-the-art techniques including the OpenCV implementations of classical detectors such as SIFT~\cite{lowe2004distinctive}, ORB~\cite{rublee2011orb}, BRISK~\cite{6126542}, and AKAZE~\cite{akaze}, as well as recent learning-based methods such as SuperPoint~\cite{detone2018superpoint}, D2-Net~\cite{dusmanu2019d2}, and RoRD~\cite{parihar2021rord}, for which we used pre-trained models obtained from the authors' GitHub repository. Note, however, that these pre-trained models were designed to perform well on perspective images, and they may not generalize well to the hybrid scenario. They are used here as baselines, given also the lack of learning-based approaches for the hybrid scenario, and to motivate the development of our approach.

We used a nearest neighbor matching strategy for all the detected descriptors within an image pair. The  Matching Score is based on a \textit{correct distance} $\epsilon$ of 3 px, 5 px, except for the Evansdale dataset, where we used a \textit{correct distance} of 5 px and 10 px due to a higher reprojection error. We resized the perspective and fisheye images to $320\times 320$ during testing, keeping a maximum of the top 300 keypoints for perspective images and the top 1000 keypoints for fisheye images to balance the information gap due to the smaller field of view of the perspective images. We use two evaluation metrics: Mean Matching Score (MMS) for descriptor evaluation, and Repeatability for detector evaluation, calculated by dividing the number of correct matches by the total matches suggested by the pipeline. We follow~\cite{detone2018superpoint} to compute the Repeatability.

Tables~\ref{tab:gtavtest-test}, \ref{tab:kittitest-test}, and \ref{tab:evansdale-test} summarize the performance of our method. It can be seen that the proposed rigorous training with random homography images, including extreme random homographies, has allowed our method to consistently achieve the highest MMS.

Traditional detection methods, such as~\cite{lowe2004distinctive, rublee2011orb, 6126542, akaze}, underperform on fisheye image feature extraction and descriptor due to the strong distortion and non-linear projection inherent in fisheye cameras, which violates the assumption of local linearity that these methods rely on. Thus, they produce comparable results on both GTAV-Hybrid and KITTI-360, where distortion is less. However, they struggle when distortion increases, as it can be seen in the Evansdale dataset. These methods are rotation invariant but not distortion invariant.

ORB has been shown to have high repeatability compared to other methods, likely due to the clustering of points that allows it to produce more repeatable matches. However, the proposed method demonstrates competitive results on Evansdale, where it achieves the best repeatability score. This score depends heavily on the quality of the learned point detector, and although we trained our network to detect a diverse range of points, the number of points detected may not be sufficient for some datasets. Therefore, increasing the number of training passes could further improve performance in this area.

Learning-based approaches, such as~\cite{detone2018superpoint, dusmanu2019d2, parihar2021rord}, are trained to work with perspective images that are not affected by distortion. They yield a good number of matches as shown in Tables~\ref{tab:gtavtest-test}, \ref{tab:kittitest-test}, \ref{tab:evansdale-test}. SuperPoint is not rotation invariant and does not perform well when faced with extreme rotation, whereas D2-Net is not trained on homographies and it cannot deal with extreme homography or viewpoint changes. RoRD, using the D2-Net approach, developed a rotation invariant version that performs better than D2-Net on these test datasets.

The proposed method, on the other hand, despite not matching the overall repeatability of ORB or SuperPoint, it generates a robust descriptor for the hybrid scenario, which is effective with both synthetic and real image datasets. This is because it is designed to be distortion and rotation invariant, making it more effective for datasets with high levels of distortion, such as Evansdale. Therefore, we infer that our method effectively overcomes the limitations of traditional and learning-based approaches that we have discussed.


%% file: tex/5-Conclusion.tex
\section{Conclusions}
\label{sec:conclusions}

We introduced a novel approach for learning an interest point detector and descriptor that is specifically designed for hybrid camera systems. It leverages a hybrid homography model and a self-supervised learning approach, along with synthetic data generation to address the lack of real-world data in this domain. We constructed two datasets, one synthetic and one real, for evaluating the methods. Furthermore, we proposed a modified contrastive loss to enhance the learning process for descriptors. Our experiments demonstrate the effectiveness of our approach on a real world hybrid dataset, further validating its potential. We believe that this work can pave the way for developing more effective keypoint detector and descriptor models for hybrid camera systems.


%% file: tex/6-Acknowledgments.tex
\section*{Acknowledgments}

This material is based upon work supported by the National Aeronautics and Space Administration under Grant No. WV-80NSSC17M0053 issued through the NASA EPSCoR Program, and also by the National Science Foundation under Grant No. 1657179.
